\documentclass[letterpaper, 10 pt, conference]{ieeeconf}  %

\usepackage{graphicx}
\usepackage{amsmath,amsfonts,amssymb,subfigure,multirow,xcolor}
\usepackage[mathscr]{eucal}

\newcommand{\figref}[1]{{Fig.~\ref{fig:#1}}}
\newcommand{\secref}[1]{{Sec.~\ref{sec:#1}}}

\newcommand{\cf}[1]{(cf.~#1)}

\newcommand{\lft}{1}%
\newcommand{\rght}{2}%
\newcommand{\torque}{12}%
\newcommand{\touchdown}{\text{touchdown}}%
\newcommand{\liftoff}{\text{liftoff}}%

\newcommand{\Tstate}{v}
\newcommand{\Tinp}{w}

\newcommand{\state}{x}

\newcommand{\inp}{u}
\newcommand{\optstate}{\xi}
\newcommand{\optinp}{\mu}

\newcommand{\State}{\e{X}}

\newcommand{\Inp}{\e{U}}
\newcommand{\ssState}{X}
\newcommand{\ssInp}{U}
\newcommand{\cost}{c}
\newcommand{\Ell}{\e{L}}
\newcommand{\val}{\nu}
\newcommand{\policy}{\pi}
\newcommand{\flow}{\phi}
\newcommand{\D}{D}
\newcommand{\C}{C}
\newcommand{\T}{T}

\newcommand{\PC}{PC}

\newcommand{\set}[1]{\left\{ #1 \right\}}
\newcommand{\paren}[1]{\left( #1 \right)}

\newcommand{\abs}[1]{\left| #1 \right|}
\newcommand{\norm}[1]{\left\| #1 \right\|}

\newcommand{\st}{\mid}

\newcommand{\eqnn}[1]{\begin{equation}\begin{aligned} #1 \end{aligned}\end{equation}}
\newcommand{\sm}{\setminus}

\newcommand{\into}{\rightarrow}

\newcommand{\R}{\mathbb{R}}

\newcommand{\N}{\mathbb{N}}
\newcommand{\tr}{\top}

\newcommand{\e}{\mathscr}

\newtheorem{proposition}{Proposition}
\newtheorem{definition}{Definition}
\newtheorem{theorem}{Theorem}
\newtheorem{corollary}{Corollary}
\newtheorem{lemma}{Lemma}
\newtheorem{claim}{Claim}
\newtheorem{assumption}{Assumption}
\newtheorem{remark}{Remark}

\newtheorem{example}{Example}

\newcommand{\thm}[1]{\begin{theorem} #1 \end{theorem}}

\usepackage{cite}

\usepackage{todonotes}

\IEEEoverridecommandlockouts                              %

\title{\LARGE \bf
Nonsmooth optimal value and policy functions 
\\
in 
mechanical systems subject to unilateral constraints
}

\author{
Bora~S.~Banjanin%
\and Samuel~A.~Burden%
\thanks{Dept.\ of Electrical \& Computer Engineering, University of Washington, Seattle, WA, USA ({\tt borab,sburden@uw.edu}).  +1~(206)~225~3545.
This material is based upon work supported by %
the U.~S.~Army Research Office 
under grant number W911NF-16-1-0158.}%
}

\begin{document}

\maketitle
\thispagestyle{empty}
\pagestyle{empty}

\begin{abstract}

State-of-the-art approaches to optimal control use 
smooth approximations of value and policy functions 
and 
gradient-based algorithms for improving approximator parameters. 
Unfortunately, we show that value and policy functions that arise in optimal control of mechanical systems subject to unilateral constraints -- i.e. the contact-rich dynamics of robot locomotion and manipulation -- are generally nonsmooth due to the underlying dynamics exhibiting discontinuous or piecewise-differentiable trajectory outcomes.
Simple mechanical systems are used to illustrate 
this result 
and
the implications 
for optimal control of contact-rich robot dynamics.

\end{abstract}

\section{Introduction}
\label{sec:intro}

This paper focuses on optimal control of \emph{mechanical} systems subject to \emph{unilateral} constraints~\cite{Ballard2000-ui}, which are commonly used to model contact-rich dynamics of rigid robots~\cite{Johnson2016-nh}.
In an optimal control problem, a policy is sought that extremizes a given performance criterion; the performance achieved by this \emph{optimal policy} is the \emph{optimal value} of the problem.
Two popular approaches for solving such problems are 
trajectory optimization~\cite{Polak1997-xd}
and 
reinforcement learning~\cite{Bertsekas1996-lf}.
Although many algorithms are available in either framework, scalable algorithms in both leverage local approximations -- gradients of values and/or policies -- to iteratively improve toward optimality.
In applications with smooth dynamics, these gradients are guaranteed to exist and can be readily computed or approximated.

Recent work has applied state-of-the-art algorithms for 
trajectory optimization~\cite{Posa2014-cz,Patel2019-nr,Mordatch2012-ar,Hereid2018-vy} 
and 
reinforcement learning~\cite{Schulman2015-kr,Kumar2016-nt,Levine2016-lw,Hwangbo2019-ou}
to optimal control of contach-rich dynamics, producing impressive results in simulations and experiments of robot manipulation and locomotion.
However, the algorithms underlying these results~\cite{Polak1997-xd,Williams1992-as} are only known to converge to stationary points in smooth systems since they rely on gradients of the functions that define costs and constraints.

\subsection{Our contributions}
We show that these gradients generally fail to exist for mechanical systems subject to unilateral constraints due to nonsmoothness in the underlying dynamics; 
this result is derived theoretically in Theorem~\ref{thm:pc} 
and demonstrated using the simple mechanical system depicted in~\figref{mdls}.
These contributions imply that additional work is required to justify applying state-of-the-art algorithms for optimal control to mechanical systems with contact-rich robot dynamics.

\subsection{Organization}
We begin in~\secref{sys} by modeling contact-rich robot dynamics using mechanical systems subject to unilateral constraints, and describe how nonsmoothness -- discontintinuity or piecewise-differentiability -- manifests in trajectory outcomes and (hence) trajectory costs. 
Then in~\secref{reg} we provide mathematical derivations that show nonsmoothness in trajectory outcomes and costs gives rise to nonsmoothness in optimal value and (hence) policy functions.
Subsequently in~\secref{opt} we present numerical simulations that demonstrate discontinuous or merely piecewise-differentiable optimal value and policy functions in a mechanical system subject to unilateral constraints. 
Finally in~\secref{disc} we discuss the prevalence of nonsmoothness in applications and how the lack of classical differentiability prevents gradient-based algorithms from converging to optimality.

\section{Contact-rich robot dynamics}
\label{sec:sys}

\begin{figure}

\centering
\subfigure[\emph{touchdown} maneuver]{%
\includegraphics[width=.2\textwidth]{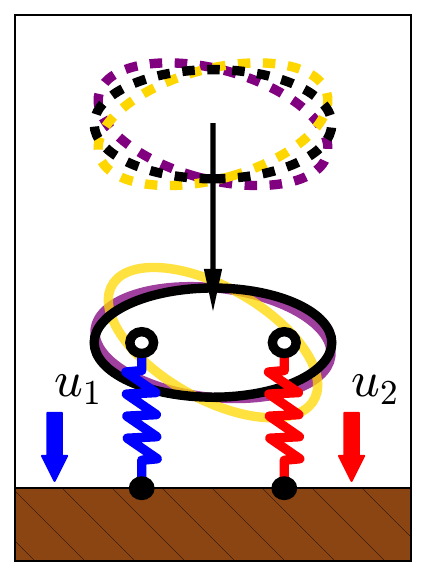}%
}
\subfigure[\emph{liftoff} maneuver]{%
\quad\includegraphics[width=.2\textwidth]{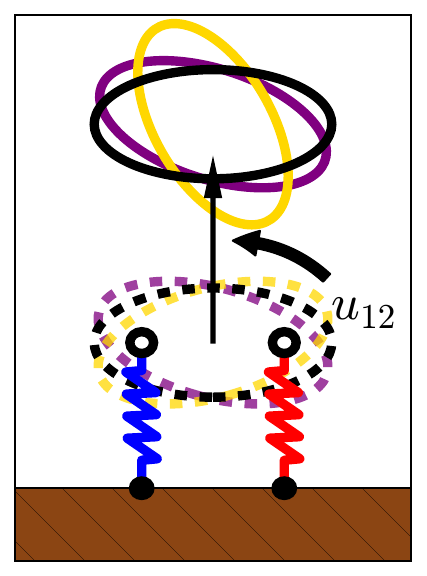}%
}%
\caption{\label{fig:mdls}
Saggital-plane biped performs two maneuvers with contact-rich dynamics -- (a) \emph{touchdown} and (b) \emph{liftoff} -- 
using policies that exert different forces %
depending on which feet are in contact with the ground.
In the \emph{touchdown} maneuver, feet are initially off the ground and trajectories terminate when the body height reaches nadir;
in the \emph{liftoff} maneuver, feet are initially on the ground and trajectories terminate when the body height reaches apex.
\figref{outcomes} shows that the final body rotation is a nonsmooth -- piecewise-differentiable or discontinuous -- function of initial body rotation.
}
\end{figure}

\begin{figure*}
\centering
\subfigure[touchdown trajectory outcomes]{
\includegraphics[width=.48\textwidth]{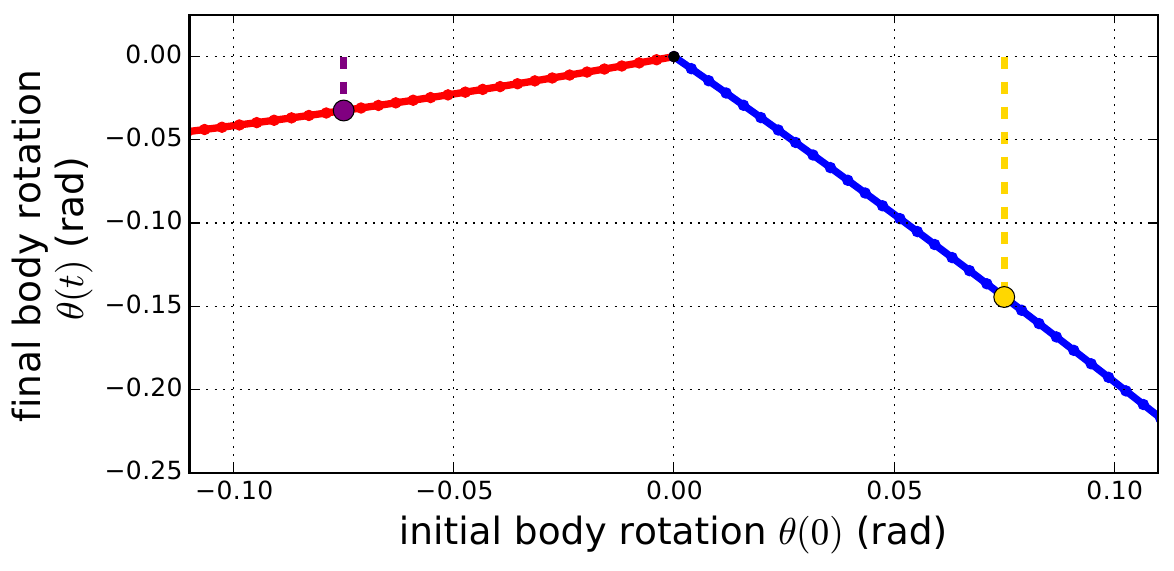}
}
\subfigure[liftoff trajectory outcomes]{
\includegraphics[width=.48\textwidth]{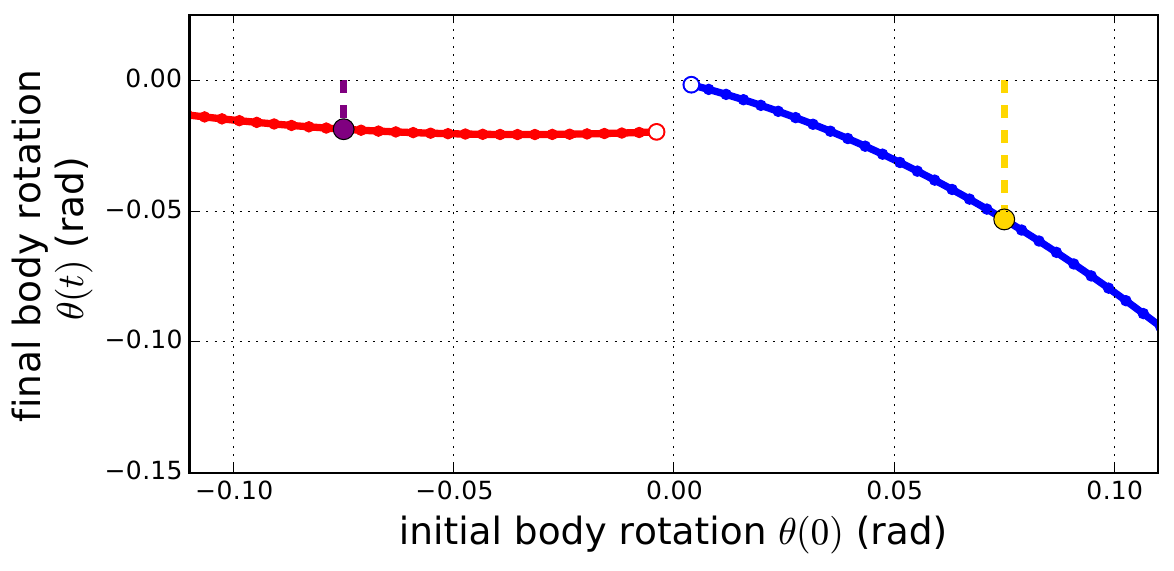}
}
\subfigure[touchdown value]{
\includegraphics[width=.48\textwidth]{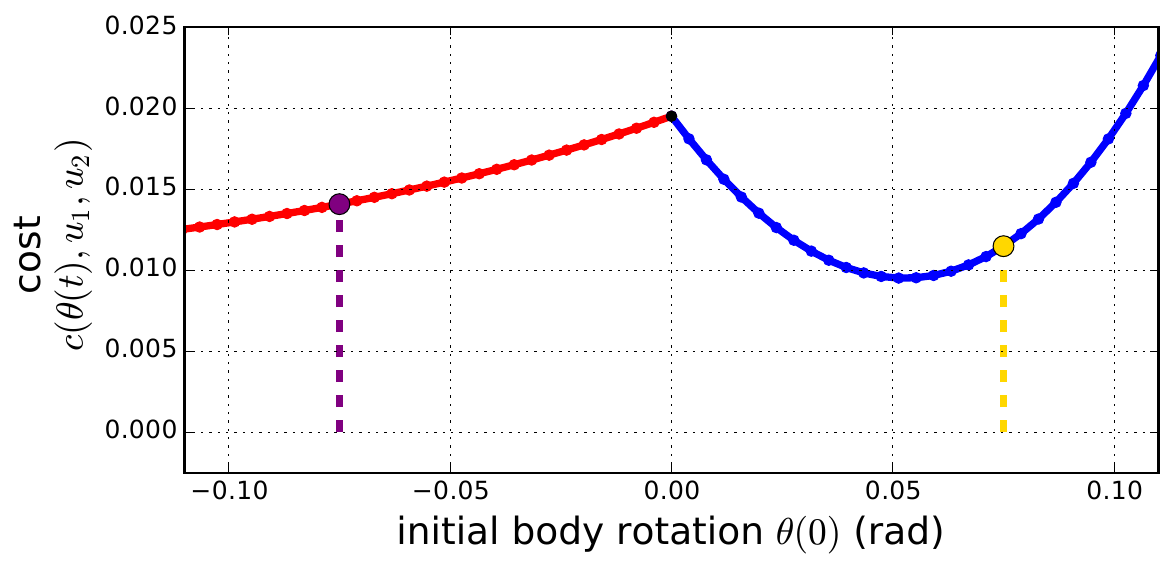}
}
\subfigure[liftoff value]{
\includegraphics[width=.48\textwidth]{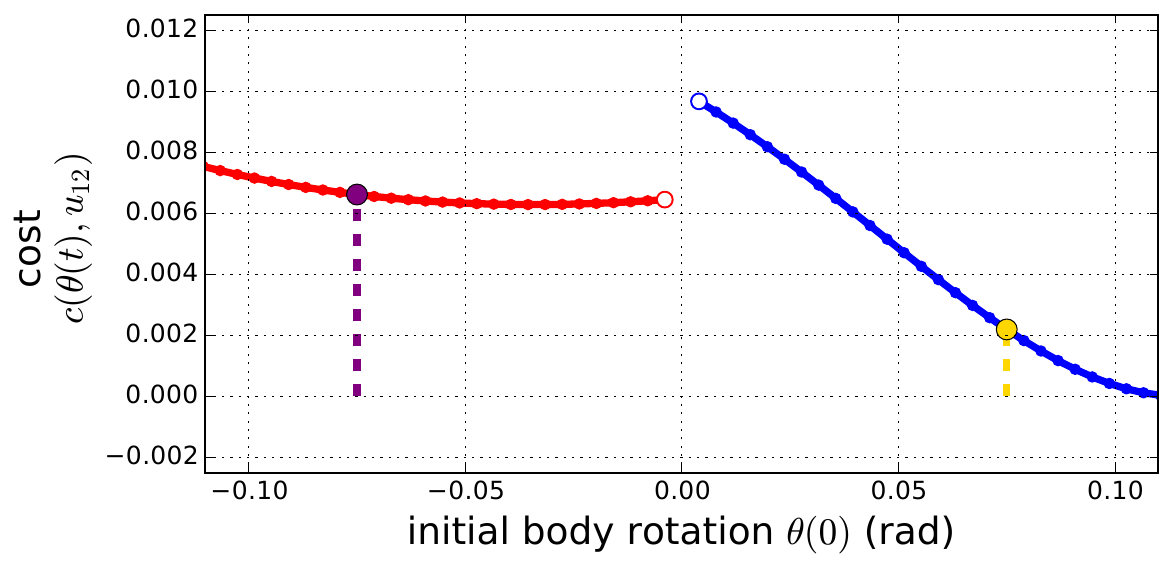}
}

\caption{\label{fig:outcomes}
\emph{Piecewise-differentiable and discontinuous trajectory outcomes in the saggital-plane biped from~\figref{mdls}.} 
(a,b) Trajectory outcomes (final body angle $\theta(t)$) as a function of initial body angle $\theta(0)$.
(c,d) Performance of trajectories as measured by the cost functions in~\eqref{eq:touchdown:cost},~\eqref{eq:liftoff:cost}.
Dashed colored vertical lines indicate corresponding colored outcomes in~\figref{mdls} and dashed lines in~\figref{contact}.
}
\end{figure*}

\begin{figure*}
\centering
\subfigure[touchdown contact modes]{
\includegraphics[width=.48\textwidth]{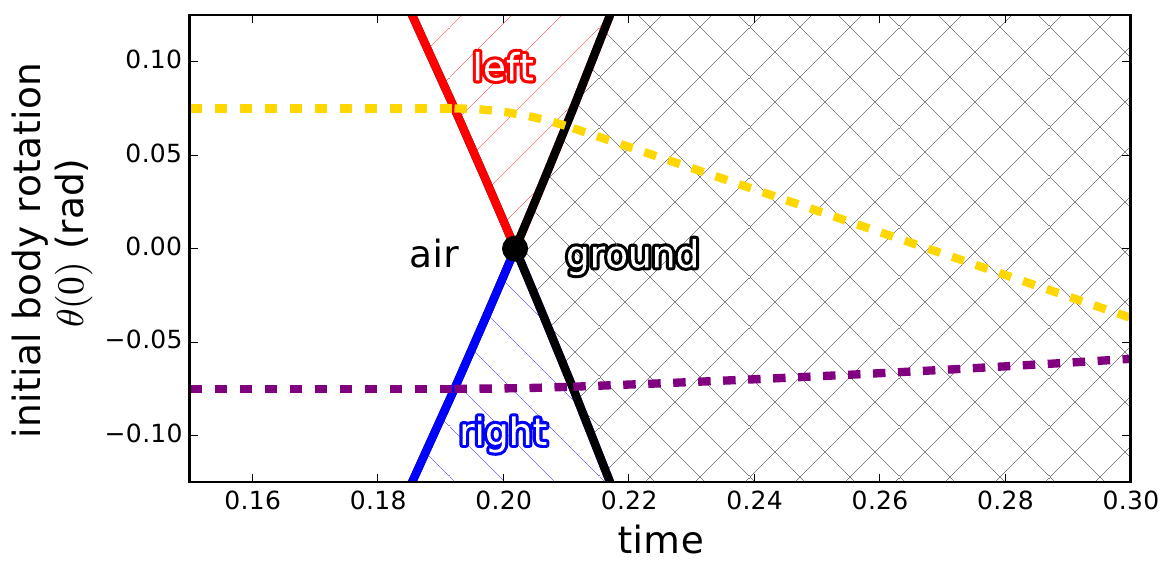}
}
\subfigure[liftoff contact modes]{
\includegraphics[width=.48\textwidth]{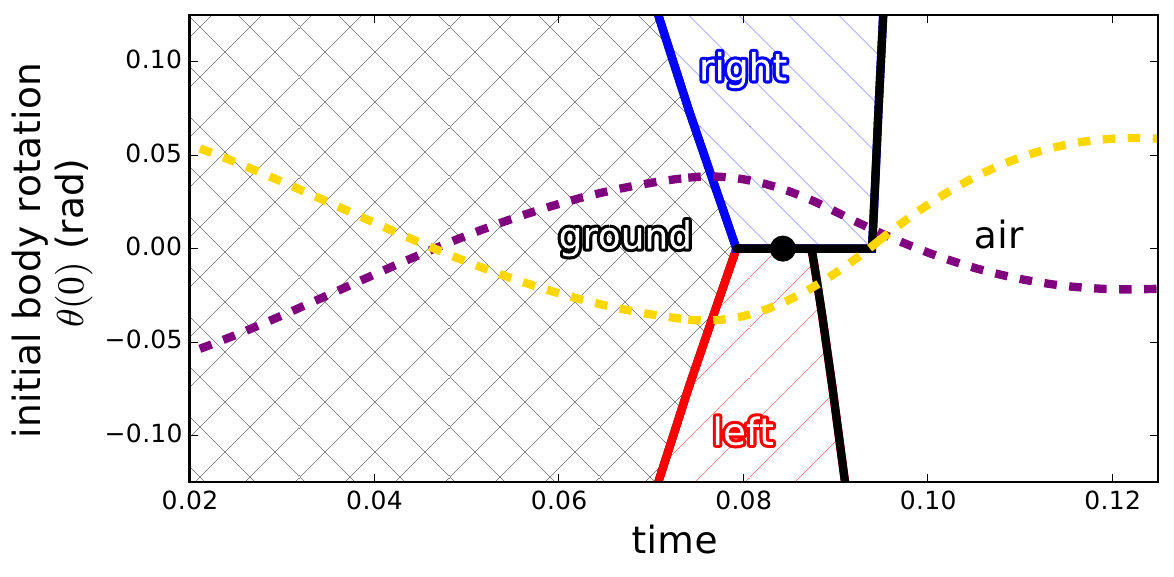}
}

\caption{\label{fig:contact}
\emph{Contact modes for touchdown and liftoff maneuvers.}
The saggital-plane biped illustrated in~\figref{mdls}(a,b) can be in one of four \emph{contact modes} corresponding to which subset $J\subset\set{1,2}$ of the (two) limbs are in contact with the ground; each subset yields different dynamics in~\eqref{eq:dyn}.
(a,b) System contact mode 
at each time $t$ 
for a given initial body rotation $\theta(0)$;
the body torque input is zero ($\inp_{12} = 0$) and the leg forces are different ($\inp_1 \ne \inp_2$) in mode \emph{left} ($\set{1}$) and \emph{right} ($\set{2}$) than in \emph{air} ($\emptyset$) or \emph{ground} ($\set{1,2}$).
Dashed colored horizontal lines indicate corresponding colored trajectories in~\figref{mdls}.
The increase in force 
during the transition to modes \emph{left} and \emph{right}
in (b) 
changes the ground reaction force discontinuously, delaying liftoff %
and causing
discontinuous trajectory outcomes in~\figref{mdls}(d).
}
\end{figure*}

In this section, we formalize a class of models for contact-rich dynamics in robot locomotion and manipulation as \emph{mechanical systems subject to unilateral constraints} 
and 
formulate an optimal control problem for these systems.

\subsection{Dynamics}
\label{sec:dyn}

Consider the dynamics of a mechanical system with 
$d\in\N$ degrees-of-freedom (DOF) $q\in Q=\R^d$ 
subject to 
$n\in\N$ unilateral constraints $a(q) \ge 0$ 
specified by a continuously-differentiable function 
$a : Q\into \R^n$,
where the inequality is enforced componentwise.
Given any $J\subset\set{1,\dots,n}$, 
and letting $\abs{J}$ denote the number of elements in the set $J$,
we let 
$a_J : Q \into \R^{\abs{J}}$ 
denote the function obtained by selecting the component functions of $a$ indexed by $J$. 
It is well-known~\cf{~\cite[Sec.~3]{Ballard2000-ui} or \cite[Sec.~2.4,~2.5]{Johnson2016-nh}}
that, with
$J = \set{j\in\set{1,\dots,n} : a_j(q) = 0}$ 
denoting the \emph{contact mode},
the system's dynamics take the form
\begin{subequations}\label{eq:dyn}
\begin{align}
  M(q)\ddot{q} & = f_J(q,\dot{q},\inp) + Da_J(q)^\tr \lambda_J(q,\dot{q},u),\label{eq:dyn:cont}\\
  \dot{q}^+ & = \Delta_J(q)\dot{q}^-,\label{eq:dyn:disc}
\end{align}
\end{subequations}
where:
$M(q)\in\R^{d\times d}$
is the mass matrix;
$f_J(q,\dot{q},u)\in\R^{d}$ is the vector of Coriolis, potential, and applied forces;
$\inp\in\Inp$ is an external input,
$Da_J(q)\in\R^{\abs{J}\times d}$ 
denotes the derivative of the constraint function $a_J$;
$\lambda_J(q,\dot{q},u)\in\R^{\abs{J}}$ 
denotes the reaction forces generated in contact mode $J$ to enforce $a_J(q) \ge 0$;
$\Delta_J(q)\in \R^{d\times d}$ 
specifies the collision restitution law that instantaneously resets velocities to ensure compatibility with the constraint $a_J(q) = 0$;
and
$\dot{q}^+$ (resp. $\dot{q}^-$) denotes the right- (resp. left-)handed limits of the velocity with respect to time.
Note that we explicitly allow the dynamics in~\eqref{eq:dyn} to vary with contact mode $J\subset\set{1,\dots,n}$.

\subsection{Properties of dynamics}
\label{sec:prop:dyn}
The seemingly benign equations in~\eqref{eq:dyn} 
can yield dynamics with a range of regularity properties.
This issue has been investigated elsewhere~\cite{Pace2017-tt, Pace2017-ph, Ballard2000-ui};
here we focus specifically on how the design of a robot's \emph{mechanical} and \emph{control} systems affect 
(non)smoothness of trajectory outcomes.

It is common to assume that the functions in~\eqref{eq:dyn} are continuously-differentiable; %
however, as illustrated by~\cite[Ex.~2]{Ballard2000-ui}, this assumption alone does not ensure even existence or uniqueness of trajectories, let alone smoothness of trajectory outcomes.
This case contrasts starkly with that of a smooth differential or difference equation
\eqnn{\label{eq:c:dyn}
\dot{x}\ \text{or}\ x^+ = F(x,u),
}
which yields unique trajectories that vary smoothly with respect to 
state 
$\state\in\State$ 
and 
control input
$\inp\in\Inp$~\cite[Thm.~5.6.8]{Polak1997-xd}. 
Since we are chiefly concerned with how properties of the dynamics in~\eqref{eq:dyn} affect properties of optimal value and policy functions, 
we will assume%
\footnote{We refer the interested reader to~\cite[Thm.~10]{Ballard2000-ui} or~\cite{Johnson2016-nh} for conditions that ensure existence and uniqueness of trajectories of~\eqref{eq:dyn}.
}
in what follows that conditions have been imposed to ensure unique trajectories of~\eqref{eq:dyn} exist for time horizons, initial states, and control inputs of interest. 

Assuming that unique trajectories exist for~\eqref{eq:dyn} does not provide any regularity properties on the trajectory outcomes;
these properties are determined by 
the design of a robot's mechanical and control systems 
and
their closed-loop interaction with the environment.
For instance: 
when limbs are inertially coupled (e.g. by rigid struts and joints), so that one limb's constraint activation instantaneously changes another's velocity, 
trajectories can vary discontinuously near configurations where these two limbs activate constraints simultaneously~\cite[Table~3]{Remy2010-vo}~\cite{Hurmuzlu1994-wk};
when limbs are force coupled (e.g. by a mode-switching controller), so that one limb's constraint (de)activation instantaneously changes the force on another, 
trajectories can vary piecewise--differentiably near configurations where these two limbs (de)activate constraints simultaneously~\cite[Fig.~1]{Pace2017-tt}.
It is this force coupling, explicitly permitted by the mode-dependence of the dynamics in~\eqref{eq:dyn}, that we will leverage to obtain nonsmooth trajectory outcomes in the examples presented in~\secref{opt}.

\subsection{Properties of optimal value and policy functions}
\label{sec:prop:opt}

A broad class of optimal control problems for the dynamics in~\eqref{eq:dyn} can be formulated in terms of
\emph{final} 
($\ell:\State\into\R$)
and 
\emph{running}
($\Ell:[0,t]\times\State\times\Inp\into\R$)
costs:
\eqnn{\label{eq:min:ctrl_}
\val(\state) = \min_{\inp\in \Inp^{[0,t]}} \ell(\flow^{\state,\inp}(t)) + \int_0^t \Ell(s,\flow^{\state,\inp}(s),\inp)\, ds,
}
where $\flow^{\state,\inp}:[0,t]\into\State$ denotes the unique trajectory obtained from initial state $\flow^{\state,\inp}(0) = \state\in\State$ when input $\inp\in\Inp$ is applied. %
To expose the dependence of the cost in~\eqref{eq:min:ctrl_} on the trajectory outcome function $\phi$, we transcribe the problem in~\eqref{eq:min:ctrl_} to a simpler form using a standard state augmentation technique~\cite[Ch.~4.1.2]{Polak1997-xd},
\eqnn{\label{eq:min:ctrl}
\val(\state) = \min_{\inp\in \Inp} \cost\paren{\flow(t,\state,\inp)},%
}
where $\flow:[0,t]\times\State\times\Inp\into\State$ is the \emph{flow} function defined by $\flow(t,\state,\inp) = \flow^{\state,\inp}(t)$.
As discussed in~\secref{prop:dyn}, the 
properties
of ${\flow}$ 
are
determined by a robot's design: %
it is possible for ${\flow}$ and hence $\cost\circ\flow$ to be 
discontinuous (${\flow}\not\in\C^0$)
or 
piecewise-differentiable and not continuously-differentiable (${\flow}\in\PC^r\sm\C^r$)
depending on the properties of the robot's mechanical and control systems.
In the next section, we study how continuity and differentiability properties of 
the cost 
$\cost\circ\flow$ 
affect the corresponding properties of 
the value 
$\val$ 
in~\eqref{eq:min:ctrl}.

\section{Nonsmooth optimal value \& policy functions}
\label{sec:reg}

Consider minimization of the {cost function} $c:\State\times\Inp\into\R$ with respect to an {input} $\inp\in \Inp$:
\eqnn{\label{eq:min}
\val(\state) = \min_{\inp\in \Inp} \cost(\state,\inp);
}
so long as $\State$ and $\Inp$ are compact and $\cost$ is continuous,
the function $\nu:\State\into \R$ indicated in~\eqref{eq:min}, termed the \emph{optimal value function}, is well-defined.
We let $\policy:\State\into \Inp$ denote an \emph{optimal policy} for~\eqref{eq:min}, i.e.
\eqnn{\label{eq:argmin}
\forall \state\in \State : \policy(\state) & \in \arg\min_{\inp\in \Inp} \cost(\state,\inp)\\ 
}
or, equivalently,
\eqnn{\label{eq:value}
\forall \state\in \State : \val(\state) & = \cost(\state,\policy(\state)).
}
In this section we study how continuity and differentiability properties of the cost function ($c$) relate to corresponding properties of the optimal value ($\val$) and policy ($\policy$) functions.

\subsection{Discontinuous cost functions}
\label{sec:value:nc}
If the cost 
($\cost:\State\times\Inp\into\R$) is discontinuous with respect to its first argument, then the optimal policy 
($\policy:\State\into\Inp$) and value 
($\val:\State\into\R$) are generally discontinuous as well.
This observation is clear in the trivial case that the cost only depends on its first argument, but manifests more generally.

\subsection{Piecewise-differentiable cost functions}
\label{sec:value:pc}

If $\cost$ is piecewise-differentiable,%
\footnote{We use the notion of piecewise-differentiability from~\cite[Ch.~4.1]{Scholtes2012-la}: a function is piecewise-differentiable if it is everywhere locally a continuous selection of a finite number of continuously-differentiable functions.}
which we denote by ${\cost\in\PC^1(\State\times \Inp,\R)}$ or simply $\cost\in\PC^1$, 
then necessarily%
\eqnn{\label{eq:pc:nec}
\forall \Tinp\in\T_\inp\Inp 
: 
\D_2 \cost(\state,\policy(\state);\Tinp) \ge 0.
}
Here, %
$\D_2 \cost(\state,\policy(\state)):\T_\inp\Inp\into\R$ 
denotes a continuous and piecewise-linear first-order approximation termed the \emph{Bouligand} (or \emph{B-})derivative~\cite[Ch.~3]{Scholtes2012-la} that exists by virtue of the cost being $\PC^1$~\cite[Lem.~4.1.3]{Scholtes2012-la}; 
$\D_2 \cost(\state,\policy(\state);\Tinp)$ 
denotes the evaluation of
$\D_2 \cost(\state,\policy(\state))$
at $\Tinp\in\T_\inp\Inp$.

If 
$\cost$ is two times piecewise-differentiable ($\cost\in\PC^2$), 
and if a sufficient condition~\cite[Thm.~1]{Chaney1988-zq} for strict local optimality%
~for~\eqref{eq:min} is satisfied at $\policy(\state)\in\Inp$,
\eqnn{\label{eq:pc:suf}
\forall \Tinp\in \set{\Tinp\in\T_\inp\Inp \st \Tinp\ne 0,\ \D_2\cost(\state,\policy(\state);\Tinp) = 0}
\\
:\D_2^2 \cost(\state,\policy(\state);\Tinp,\Tinp) > 0,
}
and if the piecewise-linear function%
\eqnn{\label{eq:pc:stab}
\ \D_2^2 \cost(\state,\policy(\state)):\T_\inp\Inp\into\T_\inp\Inp\ \text{is invertible},
}
then a $\PC^1$ Implicit Function Theorem can be applied to choose $\policy\in\PC^1$ near $\state$~\cite[Cor.~3.4]{Robinson1991-xg}. %
Applying the $\PC^1$ Chain Rule~\cite[Thm.~3.1.1]{Scholtes2012-la} to~\eqref{eq:pc:nec} yields~\cf{~\cite[\S~3]{Robinson1991-xg}}
\eqnn{\label{eq:pc:Dpolicy}
\forall \Tstate\in&\T_\state\State : \D\policy(\state;\Tstate) = \\
&- \D_2^2 \cost(\state,\policy(\state))^{-1}\paren{\D_{12}\cost(\state,\policy(\state);\Tstate)},
}
and applying the $\PC^1$ Chain Rule to~\eqref{eq:value} yields
\eqnn{\label{eq:pc:Dval}
\forall \Tstate\in&\T_\state\State : \D\val(\state;\Tstate) = \D_\state \cost(\state,\policy(\state);v) \\
& = \D_1 \cost(\state,\policy(\state);\Tstate) + \D_2 \cost(\state,\policy(\state);\D\policy(\state;\Tstate)),
}
whence we obtain B-derivatives of the optimal value and policy functions in terms of B-derivatives of the cost.

We conclude that if the cost function is two times piecewise-differentiable ($\cost\in\PC^2$) and first-order necessary~\eqref{eq:pc:nec} and second-order sufficient~\eqref{eq:pc:suf},~\eqref{eq:pc:stab} conditions for optimality and stability of solutions to~\eqref{eq:min} are satisfied at $\inp = \policy(\state)$, then the optimal policy and value functions are piecewise-differentiable at $\state$ ($\policy,\val\in\PC^1$) and their B-derivatives at $\state$ can be computed using~\eqref{eq:pc:Dpolicy},~\eqref{eq:pc:Dval}.

\thm{\label{thm:pc}
If 
$\cost\in\PC^2(\State\times\Inp,\R)$ satisfies~\eqref{eq:pc:nec},~\eqref{eq:pc:suf}, and~\eqref{eq:pc:stab} at $(\optstate,\optinp)\in\State\times\Inp$,
then 
there exist neighborhoods
$\ssState\subset\State$ of $\optstate$ 
and
$\ssInp\subset\Inp$ of $\optinp$ 
and
a function
$\policy\in\PC^1(\ssState,\ssInp)$ 
such that $\policy(\optstate) = \optinp$ 
and,
for all $\state\in\ssState$,
$\policy(\state)$ is the unique minimizer for
\eqnn{
\val(x) = \min_{\inp\in\ssInp} \cost(\state,\inp);
}
the B-derivative of $\policy$ is given by~\eqref{eq:pc:Dpolicy}, 
and
the B-derivative of $\val$ is given by~\eqref{eq:pc:Dval}.
}

\subsection{Conclusions about optimal value \& policy functions}
The results in Sections~\ref{sec:value:nc} %
and~\ref{sec:value:pc} suggest that we should generally expect continuity and differentiability properties of optimal value and policy functions to match that of the cost function:  they should be discontinuous when the cost is discontinuous, or piecewise-differentiable when the cost is piecewise-differentiable.
In~\secref{opt} we demonstrate these effects in the class of models described in~\secref{sys}. %

\section{Nonsmooth optimal value \& policy functions in contact-rich robot dynamics}
\label{sec:opt}

We showed in the previous two sections that optimal value and policy functions 
inherit nonsmoothness from the underlying dynamics.
To instantiate this result, we crafted a simple mechanical system subject to unilateral constraints that exhibits 
piecewise-differentiable and discontinuous trajectory outcomes,
yielding the \emph{touchdown} and \emph{liftoff} maneuvers shown in~\figref{mdls}(a,b).
For the touchdown maneuver, we seek the optimal (constant) force to exert in the left leg ($\inp_{\lft}$) when the left foot is in contact and the right foot is not; similarly, we seek the optimal choice of force in the right leg ($\inp_{\rght}$) when the right foot is in contact and the left foot is not: 
with 
$\theta^*$ denoting the desired body rotation at nadir 
and 
$\alpha_{\lft},\alpha_{\rght} > 0$ denoting input penalty parameters,
\eqnn{\label{eq:touchdown:cost}
\cost_\touchdown(\theta,\inp_{\lft},\inp_{\rght})=(\theta-\theta^*)^2 + \alpha_{\lft} \inp_{\lft}^2 + \alpha_{\rght} \inp_{\rght}^2.
}
For the liftoff maneuver, we seek the optimal (constant) torque ($\inp_{\torque}$) to apply to the body while both feet are in contact: 
with 
$\theta^*$ denoting the desired body rotation at apex
and
$\alpha_{\torque} > 0$ denoting an input penalty parameter,
\eqnn{\label{eq:liftoff:cost}
\cost_{\liftoff}(\theta,\inp_{\torque})=(\theta-\theta^*)^2 + \alpha_{\torque} \inp_{\torque}^2.
}
We implemented numerical simulations of these models%
\footnote{using the modeling framework in~\cite{Johnson2016-nh} and simulation algorithm in~\cite{Burden2015-ip}}
and applied a scalar minimization algorithm%
\footnote{{\tt SciPy v0.19.0 minimize\_scalar}}
to compute optimal policies as a function of initial body rotation. %

\begin{figure*}
\centering
\subfigure[optimal touchdown trajectory outcomes]{
\includegraphics[width=.48\textwidth]{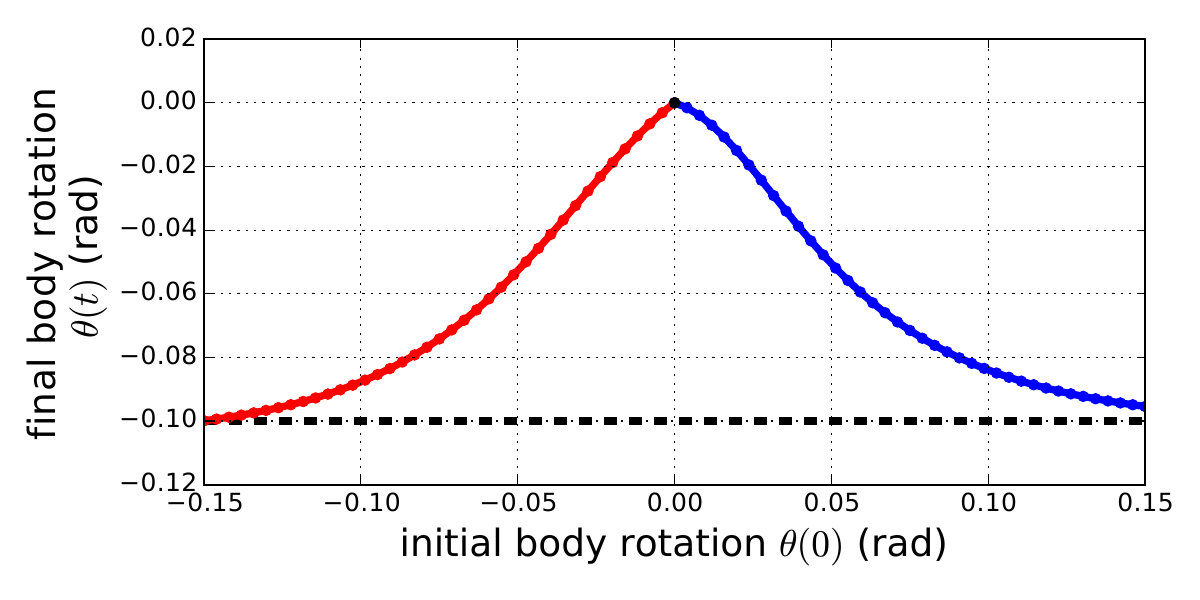}
}
\subfigure[optimal liftoff trajectory outcomes]{
\includegraphics[width=.48\textwidth]{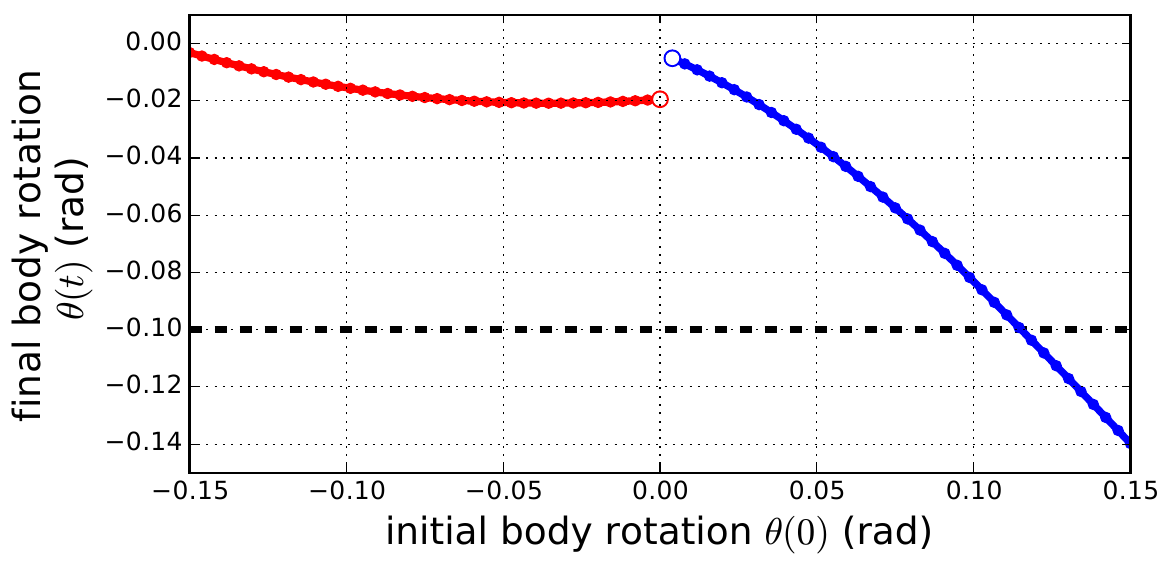}
}
\subfigure[optimal touchdown policy]{
\includegraphics[width=.48\textwidth]{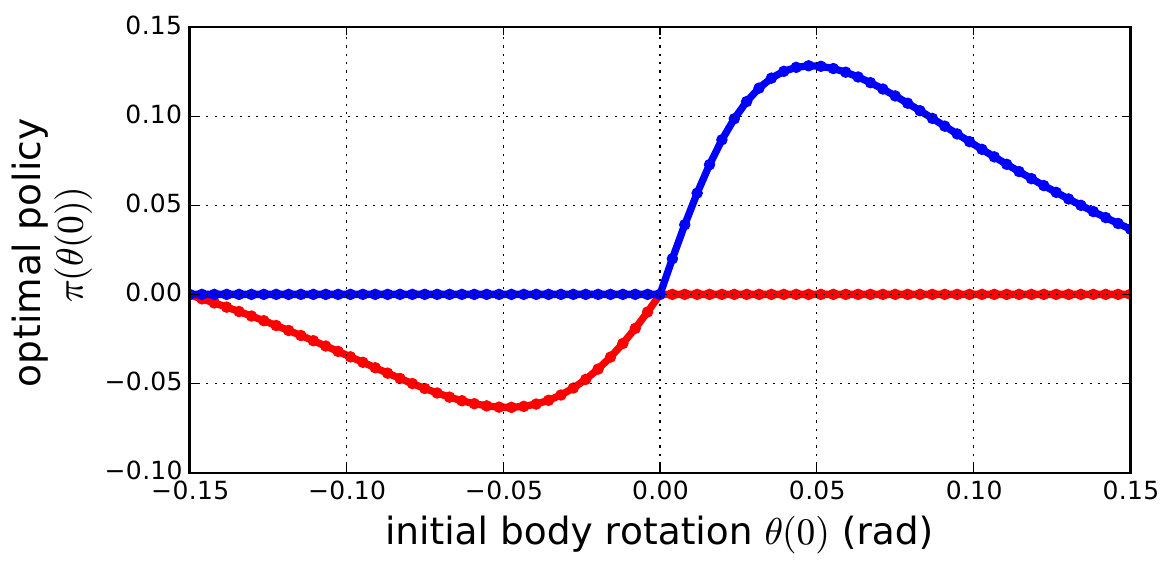}
}
\subfigure[optimal liftoff policy]{
\includegraphics[width=.48\textwidth]{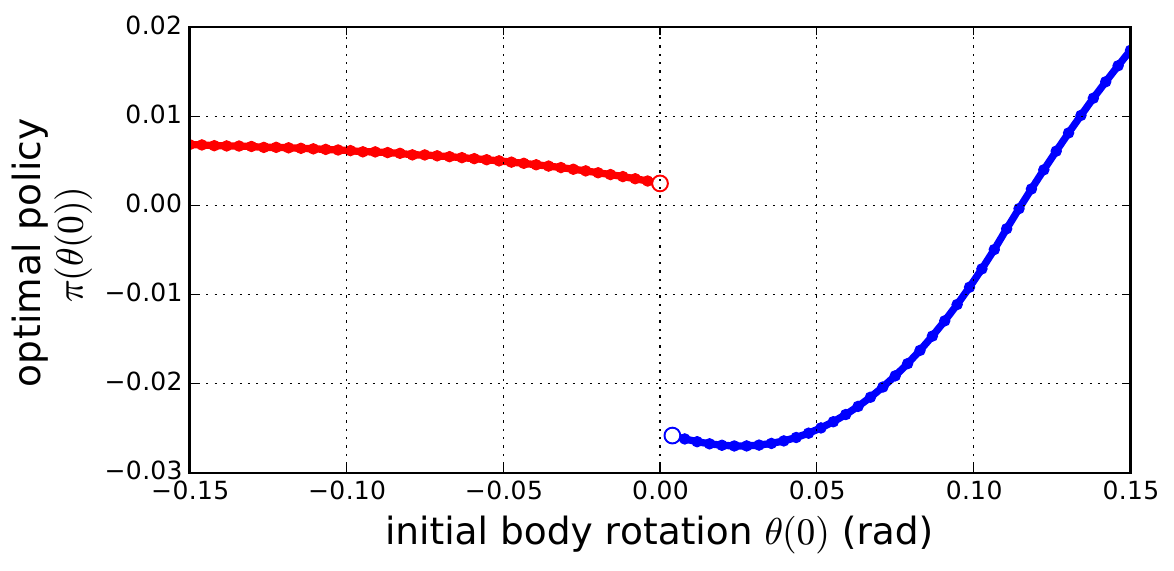}
}
\subfigure[optimal touchdown value]{
\includegraphics[width=.48\textwidth]{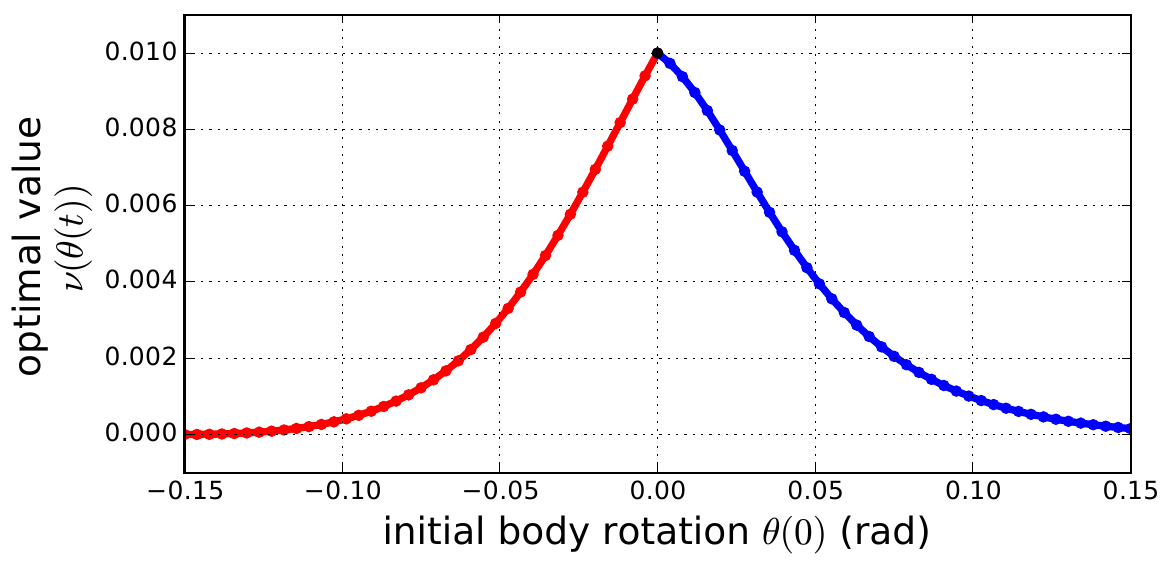}
}
\subfigure[optimal liftoff value]{
\includegraphics[width=.48\textwidth]{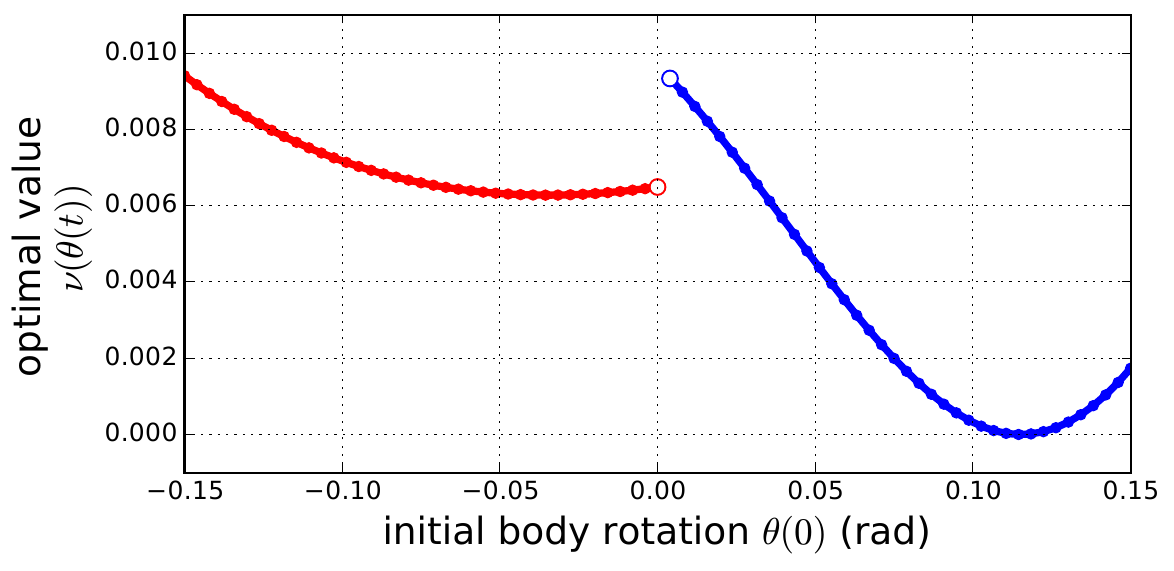}
}
\caption{\label{fig:opt}
\emph{Optimal trajectories, values and policies for touchdown and liftoff maneuvers.}
Optimizing~\eqref{eq:touchdown:cost},~\eqref{eq:liftoff:cost} for the biped in~\figref{mdls} yields trajectory outcomes (a,b), policies (c,d), and values (e,f) that are nonsmooth -- piecewise-differentiable (\emph{left}) or discontinuous (\emph{right}).
Asymmetries in trajectory outcomes are due to unequal input penalty parameters ($\alpha_1 \ne \alpha_2$) in (a) and unequal leg forces ($\inp_1 \ne \inp_2$) in (b).
}
\end{figure*}

As expected, the optimal value and policy functions computed for the touchdown and liftoff maneuvers are nonsmooth (\figref{opt}(c,d,e,f)),
owing to the nonsmoothness of the optimal trajectory outcomes (\figref{opt}(a,b)).
This result does not depend sensitively on the problem data;
nonsmoothness is preserved after altering parameters of the model and/or cost functions.
We emphasize that the nonsmoothness in~\figref{opt} arises from the nonsmoothness in the underlying system dynamics~\eqref{eq:dyn}, as the functions in~\eqref{eq:touchdown:cost} and~\eqref{eq:liftoff:cost} are smooth.

\section{Discussion}
\label{sec:disc}

We conclude by discussing
what our results imply about the use of smooth tools in nonsmooth settings (\secref{disc:smooth})
and
how often we expect to encounter the nonsmooth phenomena described above in models of robot behaviors (\secref{disc:models}).

\subsection{Justifying the use of gradient-based algorithms}
\label{sec:disc:smooth}

Suppose a (possibly non-optimal) policy 
$\policy:\State\into\Inp$ 
has an associated value
$\val^\policy:\State\into\R$. %
If this value admits a first-order approximation with respect to $\policy$,
then it is natural%
~to improve the policy by descending the cost landscape:
with $\alpha > 0$ as a stepsize parameter,
\eqnn{\label{eq:algo:policy}
\policy^+ = \policy + \alpha \arg\min_{\norm{\delta} = 1} \D_\policy\val^\policy(\delta).
}
The update in~\eqref{eq:algo:policy} is a \emph{direct policy gradient-based} algorithm~\cite{Sutton2000-ap, Baxter2001-ua}, and can be interpreted as a \emph{natural}~\cite{Kakade2001-az} or \emph{trust region}~\cite{Schulman2015-kr} algorithm depending on the norm chosen.
In practice, the derivative $\D_\policy\val^\policy$ is not known in closed-form and must be estimated, e.g. using function approximation~\cite{Doya2000-gk, Konda2003-av} or sampling~\cite{Baxter2001-ua, Silver2014-lj}.
This practice is justified for smooth control systems; %
it is not generally justified for the mechanical systems subject to unilateral constraints considered here since the value of (optimal or non-optimal) policies can be nonsmooth.

Recent work employs smooth approximations of the contact-rich robot dynamics in~\eqref{eq:dyn} to enable application of gradient-based learning~\cite{Levine2014-im, Levine2016-lw, Kumar2016-nt} and optimization~\cite{Erez2012-yo, Mordatch2012-ar, Mordatch2015-jb} algorithms.
This approach leverages established scalable algorithms,
but
does not ensure that policies optimized for the smoothed dynamics are \mbox{(near-)optimal} when applied to the original system's nonsmooth dynamics, since the dynamics of the smooth system being optimized differ from those of the original system.
As an alternative approach, 
the framework we introduced in~\cite{Pace2017-ph}
provides design conditions that ensure trajectories of~\eqref{eq:dyn} depend continuously-differentiably on initial conditions.
Thus in future work it may be possible to justify applying state-of-the-art algorithms for optimal control directly on some mechanical systems subject to unilateral constraints.

\subsection{Prevalence of nonsmoothness in contact-rich dynamics}%
\label{sec:disc:models}
In~\secref{opt}, we presented two optimal control problems where the dynamics of a mechanical system subject to unilateral constraints gave rise to a nonsmooth cost:
one where the cost was piecewise-differentiable,
and
another where it was discontinuous.
The reader may have noticed that the nonsmoothness occurred along trajectories that underwent simultaneous constraint (de)activation.
This peculiarity was not accidental:  
in the absence of dry friction, 
the cost is generally continuously-differentiable
along trajectories that (de)activate constraints at distinct time instants~\cite{Aizerman1958-ih}. %

If the constraint surfaces 
intersect transversely~\cite[Ch.~6]{Lee2012-mb},
then the nonsmoothness presented in~\secref{opt} is confined to a subset of the state space with zero Lebesgue measure.%
~In light of this observation, intuition may lead one to ignore these states in practice. %
However, we believe this intuition will lead the practitioner astray as the complexity of considered behaviors increases.
Indeed, since the number of contact mode sequences increases 
factorially with the number of constraints
and 
exponentially with the number of constraint (de)activations,
then the region where the cost function is continuously-differentiable is ``carved up'' into a rapidly increasing number of disjoint ``pieces''
as behavioral complexity %
increases.

Although we cannot at present comment in general on how these smooth pieces fit together, we note that some important behaviors will reside near a large number of pieces.
For instance, periodic behaviors with (near-)simultaneous (de)activation of $n\in\N$ constraints as in~\cite{Alexander1984-ld} could yield up to $(n!)^k$ pieces after $k\in\N$ periods~\cite[Ch.~6]{Banjanin2019-zw}.
The combinatorics are similar for tasks that involve intermittently activating (a subset of) $n$ constraints $k$ times as in~\cite{Mordatch2012-ar}.
Since the dimension of the state space is independent of $n$ and $k$, these pieces must be increasingly tightly packed as $n$ and/or $k$ increase.

Beyond the nonsmoothness induced by simultaneous constraint (de)activation considered here, mechanical systems subject to unilateral constraints can exhibit discontinuous or piecewise-differential trajectory outcomes due to dry friction, grazing, and bifurcations~\cite{Nordmark1997-ps, Bernardo2008-wh, Makarenkov2012-dk}.
Such phenomena have been studied extensively from the perspective of nonsmooth dynamics and mechanics,
but the implications for control have received comparatively little attention.
It is our hope that the results presented herein will stimulate 
interest in 
this important application domain
within the control community.

\bibliography{refs}  %

\end{document}